\documentclass[letterpaper, 10 pt, conference]{ieeeconf}
\usepackage{multirow}
\IEEEoverridecommandlockouts
\usepackage{amsmath, amssymb}
\usepackage{graphicx}
\usepackage{cite}
\usepackage{hyperref}
\usepackage{float} 
\usepackage{placeins} 
\usepackage{pifont}
\newcommand{\xmark}{\ding{55}}
\usepackage{listings}
\usepackage{xcolor}
\usepackage{placeins}
\usepackage{booktabs}
\usepackage[most]{tcolorbox}  

\lstdefinelanguage{json}{
    basicstyle=\ttfamily\footnotesize,
    stringstyle=\color{blue},
    numbers=left,
    numberstyle=\tiny\color{gray},
    stepnumber=1,
    numbersep=5pt,
    frame=none,
    backgroundcolor=\color{gray!10},
    showstringspaces=false
}
\title{\LARGE \bf
Graphormer-Guided Task Planning: Beyond Static Rules with LLM Safety Perception
}

\author{Wanjing Huang$^{1}$
, Tongjie Pan$^{2}$,  Yalan Ye$^{2}$
\thanks{$^{1}$Wanjing Huang is with the Department of Computer Science, University of California, Santa Barbara. }
\thanks{$^{2}$Yalan Ye and Tongjie Pan are with the School of Computer Science and Engineering, University of Electronic Science and Technology of China, Chengdu, Sichuan, China. Corresponding email:
{\tt\small  yalanye@uestc.edu.cn}}
}
\begin{document}

\maketitle
\thispagestyle{empty}
\pagestyle{empty}

\begin{abstract}
Recent advancements in large language models (LLMs) have  expanded their role in robotic task planning. However, while LLMs have been explored for generating feasible task sequences, their ability to ensure safe task execution remains underdeveloped. Existing methods struggle with structured risk perception, making them inadequate for safety-critical applications where low-latency hazard adaptation is required. To address this limitation, we propose a Graphormer-enhanced risk-aware task planning framework that combines LLM-based decision-making with structured safety modeling. Our approach constructs a dynamic spatio-semantic safety graph, capturing spatial and contextual risk factors to enable online hazard detection and adaptive task refinement. Unlike existing methods that rely on predefined safety constraints, our framework introduces a context-aware risk perception module that continuously refines safety predictions based on real-time task execution. This enables a more flexible and scalable approach to robotic planning, allowing for adaptive safety compliance beyond static rules.
To validate our framework, we conduct experiments in the AI2-THOR environment. The experiments results validates improvements in risk detection accuracy, rising safety notice, and task adaptability of our framework in continuous environments compared to static rule-based and LLM-only baselines. Our project is available at https://github.com/hwj20/GGTP

\end{abstract}

\section{INTRODUCTION}
 % Ensuring safety in robotic task execution is critical, especially in human-shared environments \cite{Albus1994}.
 A household robot retrieving vegetables from the refrigerator may entirely fail to notice a child approaching a hot stove. Existing robotic task planning frameworks, including rule-based systems and large language model (LLM)-driven planners, focus exclusively on predefined task sequences while lacking lwo-latency environmental risk perception \cite{pluginsafety}. These systems evaluate task execution based on internal constraints, ensuring rule compliance but ignoring external hazards. For instance, a rule-based model may enforce that a knife is only used on a cutting board but fails to recognize a child reaching for it as a risk. applications. While LLMs enable flexible, generalizable task reasoning \cite{survey1,Kannan2023}, they lack structured risk perception, making them unsuitable for safety-critical Reinforcement learning (RL) approaches \cite{RL} can optimize decision-making in controlled settings but require extensive training and fail to generalize in low-latency scenarios \cite{survey1}. We summarize three key limiations in current approaches to LLM-driven robotic task planning\cite{survey1,survey2}:

\begin{itemize}
    \item Lack of structured risk perception: Most LLM-based planners operate purely on semantic reasoning, failing to incorporate spatially structured risk modeling. As a result, they struggle to detect high-risk interactions, such as a child approaching a hazardous object.
    \item Static rule-based safety mechanisms: Traditional rule-based safety approaches require manually predefined constraints, making them rigid and difficult to scale. 
    % These methods lack adaptability to dynamically changing environments where new hazards can arise unexpectedly.
    \item Limited task adaptability: Existing frameworks typically generate task sequences in a feedforward manner, with limited capacity to modify execution plans in response to safety emergencies.
\end{itemize}

To address these challenges, we propose a Graphormer-enhanced risk-aware task planning framework that integrates structured safety perception with LLM-driven decision making. We compare with the static rule-based method in Fig. \ref{fig:overview}. Our work is inspired by the paradigm shift introduced by graph neural networks (GNNs) in autonomous driving, where structured reasoning has proven instrumental in risk prediction and trajectory optimization \cite{gnntrajectory,GNN2, gnn3,wayformer}. GNN architectures have been extensively applied to model spatial-temporal interactions, facilitating predictive control by encoding agent-object dependencies within a graph-based framework. These approaches\cite{wayformer, gnn3} employ hierarchical attention mechanisms to selectively prioritize critical relational structures, thereby refining motion forecasting and enhancing decision robustness in unstructured scenarios.

Building upon the above principles, our approach adaptively constructs a spatio-semantic safety graph, encoding both spatial and contextual risk attributes to elevate real-time hazard perception. The Graphormer\cite{graphormer} model is designed to emphasize high-risk interactions through an attention-weighted graph representation, enabling continuous risk assessment and adaptive task refinement. During execution, the system iteratively updates its risk evaluation, ensuring that, upon the emergence of a hazardous condition, a task replanning mechanism is invoked. Using LLM-based semantic reasoning in conjunction with structured risk modeling, our framework synthesizes online task adaptation with proactive safety measures, significantly improving robustness in open-world robotic applications.

\begin{figure*}[t]
    \centering
    \includegraphics[width=0.9\textwidth]{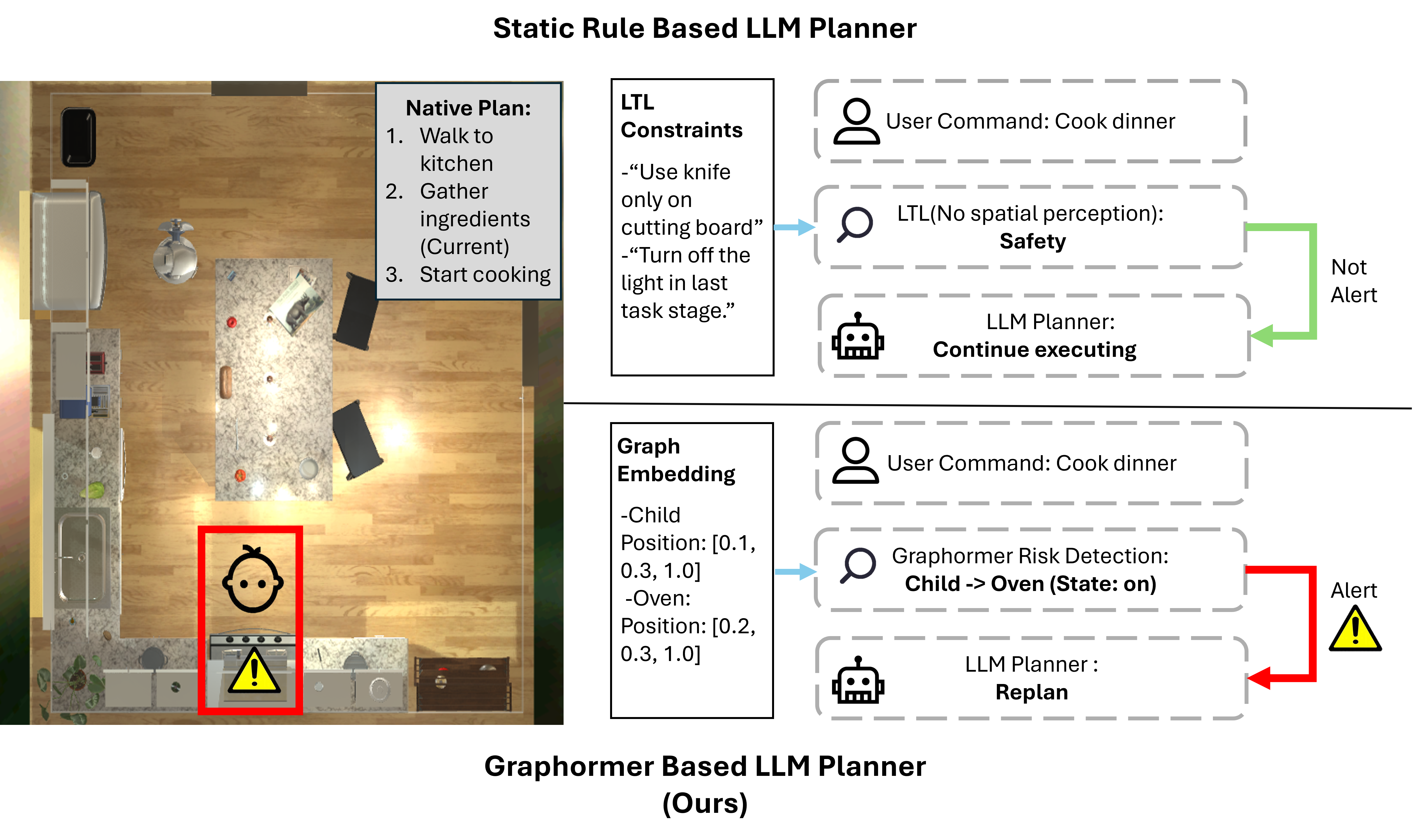}
\caption{Comparison between the static rule-based method and our approach. LTL-based safety enforcement relies on predefined constraints, which inherently lack spatial perception and fail to capture unenumerated risks. As a result, when a robot executes a task such as retrieving ingredients from the refrigerator, it is unable to recognize a child approaching a high-temperature oven. Consequently, LTL transmits an incorrect "safe" signal to the LLM planner, leading to a failure in proactive hazard avoidance. In contrast, our method models spatial-semantic relationships, enabling real-time risk assessment and adaptive task modification.}
\vspace{-10pt}
    \label{fig:overview}
\end{figure*}

% In summary, the main contributions of this study can be outlined as follows.
% \begin{itemize}

%     \item Unlike existing methods that rely on predefined safety constraints, our framework introduces a spatio-semantic safety graph to capture spatial and contextual risk factors and a context-aware risk perception module to continuously refine safety predictions based on real-time task execution. This enables a more flexible and scalable approach to robotic planning, allowing for adaptive safety compliance beyond static rules.
%     \item Extensive experiments conducted in AI2-THOR have demonstrated that our proposed framework provides substantial improvements in risk detection and adaptive task execution compared to static rule-based and LLM-only baselines.
% \end{itemize}

In summary, the main contributions of this study can be outlined as follows:
\begin{itemize}
    % \item We propose a Graphormer-enhanced risk-aware task planning framework enables adaptive task modifications when detecting risks, with flexible safety compliance for continuous environments.
    \item We introduce a spatio-semantic safety graph that models environmental risk factors based on Graphormer, providing a structured representation of hazards beyond conventional rule-based methods.
    \item We develop an adaptive  LLM decision framework that enables adaptive task modifications when detecting risks, with flexible safety compliance for continuous environments.
    \item We validate our framework through extensive experiments in AI2-THOR \cite{ai2thor}, demonstrating substantial improvements in risk detection and adaptive task execution compared to static rule-based and LLM-only baselines.
\end{itemize}

\section{RELATED WORK}

\begin{table*}[t]
    \centering
    \renewcommand{\arraystretch}{1.3} % 调整行距
    \setlength{\tabcolsep}{8pt} % 调整列间距
    \caption{Comparison of Related Work and Our Approach}
    \begin{tabular}{l c c c c}
        \toprule
        \textbf{Method} & \textbf{LLM Planning} & \textbf{Safety Model} & \textbf{Adaptivity} & \textbf{Real-time Risk Perception}\\
        \midrule
        AiGem  \cite{gnntrajectory} & \xmark & Graph-based & \checkmark & \checkmark \\
        SMART-LLM \cite{Kannan2023} & \checkmark & \xmark & \xmark & \xmark \\
        Plug in the Safety Chip \cite{pluginsafety} & \checkmark & Rule-based & \xmark & \xmark \\
        SafePlanner \cite{safetyplanner} & \checkmark & Pretrained Safety Prediction Model& \checkmark & \xmark \\
        RoCo \cite{roco} & \checkmark & \xmark & \checkmark & \xmark \\
        Co-NavGPT \cite{conavgpt} & \checkmark & \xmark & \checkmark & \xmark \\
        TidyBot \cite{tidybot} & \checkmark & \xmark & \xmark & \xmark \\
        AutoRL \cite{autorl} & \checkmark & Human Guardrails & \checkmark & \xmark \\
        Cross-Layer Sequence Supervision \cite{ltl2} & \checkmark & LTL Constraints & \xmark & \xmark \\
        \textbf{Ours} & \checkmark & Graphormer-enhanced & \checkmark & \checkmark \\
        \bottomrule
    \end{tabular}
    \vspace{-5pt}
    \label{tab:comparison}
\end{table*}

\subsection{LLMs in Robotic Task Planning}
% LLMs have been integrated into robotic systems to enhance task planning and decision-making capabilities. 
Recently rich literature has tried to employ LLMs and to accomplish comprehensive planning in complex long-horizon robotics tasks\cite{horizonplan}. Studies \cite{survey1, survey2} have shown that LLMs can generate high-level action sequences with world knowledge and semantic reasoning. LLM-driven agents have been applied to domestic robotics \cite{tidybot}, multi-agent collaboration \cite{Kannan2023,horizonplan}, and interactive task execution \cite{interactive_llm}. 
% However, while LLMs excel in natural language comprehension, their lack of structured risk perception and inability to dynamically adapt to real-time environmental hazards limit their suitability for safety-critical applications.

Despite their remarkable capabilities in task reasoning and generalization\cite{gptreport}, LLMs exhibit several fundamental challenges when applied to robotic planning and control. First, LLM-generated task sequences often suffer from inconsistencies and unpredictability, as minor variations in input prompts can lead to significantly different outputs \cite{survey2}. This lack of determinism raises critical safety concerns, particularly in real-world robotic deployments where failure to adhere to strict operational constraints can result in hazardous consequences.

Second, existing LLM-based planning frameworks rely heavily on prompt engineering, yet there is no standardized methodology for designing prompts that ensure robust and contextually aware decision-making \cite{survey2, promptengineering}. Without a structured framework for encoding safety constraints, robots may misinterpret task objectives, leading to unsafe execution behaviors.

% These limitations highlight the necessity of integrating structured safety perception with LLM-based decision-making. 
% Unlike prior approaches that solely rely on heuristic filtering mechanisms, our framework explicitly models risk at a structural level, incorporating graph-based spatio-semantic reasoning to enable real-time hazard adaptation.

\subsection{Rule-Based Safety Constraints in Robotics}
Linear Temporal Logic (LTL)  has found utility innforcing formal safety constraints in robotic planning and control \cite{ltl1, ltl2, ltl3}. Due to its expressivity and mathematically rigorous semantics, LTL provides a structured approach to specifying task constraints over time, ensuring compliance with pre-defined operational safety rules \cite{ltl4, ltl5}. LTL-based safety enforcement has been extensively explored in runtime verification and monitoring, where predefined temporal logic specifications are continuously checked against system execution traces to prevent unsafe behaviors \cite{ltl6, ltl7}. 

In industrial robotics, LTL has been adopted for programmable logic controllers (PLCs) to ensure compliance with safety standards such as ISO 61508 \cite{iso}. LTL has also been applied in human-robot interaction (HRI), where safety constraints are viewed as language-specified conditions that must be satisfied for seamless collaboration between humans and robots \cite{ltl9, ltl10}. 

However, while LTL provides strong formal guarantees, it suffers from poor adaptability to continuous environments. LTL-based methods rely on predefined logical constraints\cite{pluginsafety}, which require exhaustive enumeration of all possible safety conditions . This rigidity makes LTL-based approaches impractical for open-world robotic applications, where environmental hazards can emerge unpredictably. 
% Unlike LTL-based methods that enforce static rules, our framework integrates adaptive risk perception, allowing real-time task modification based on evolving environmental hazards.

\subsection{Safety Perception in Robotics}
Recent studies have explored integrating vision-language models (VLMs) with LLM-based robotic planning to enhance environmental awareness \cite{autorl,safetyplanner}. While such methods improve perceptual understanding, they primarily focus on task feasibility rather than structured safety reasoning. Furthermore, these approaches rely on human-in-the-loop supervision to ensure compliance, rather than enabling autonomous risk adaptation\cite{autorl}.

In autonomous systems, graph-based models have been adopted  for structured spatial data processing and risk modeling. GNNs have been used for trajectory prediction \cite{gnntrajectory}, pedestrian intent estimation \cite{gnn_pedestrian}, and interaction-aware motion planning \cite{gnninteraction} in autonomous driving. Recent studies have also explored graph-based representations for robotic navigation and spatial reasoning \cite{gnn_robot1, gnn_robot2}, enabling agents to model complex object-agent interactions. However, the integration of graph-based risk perception with LLM-driven task planning remains underexplored.

% \subsection{Comparison with Existing Methods}
Our work builds upon these advancements by integrating LLM-based semantic reasoning with Graphormer-enhanced risk-aware task adaptation. Unlike prior approaches that either rely solely on LLMs for planning or employ static rule-based safety constraints, our framework constructs a spatio-semantic safety graph that continuously updates during task execution, ensuring immediate hazard detection and proactive task modifications.

To highlight the limitations of existing LLM-driven task planning and rule-based safety mechanisms, we summarize key differences in Table \ref{tab:comparison}.

\section{METHODOLOGY}

\subsection{Problem Formulation}
Robotic agents operating in open-world environments must balance task efficiency with operational safety \cite{Albus1994}. We propose a framework that integrates a spatio-semantic safety graph with an LLM-driven task planner to achieve risk-aware task execution. Our method constructs a instantaneous risk representation of the environment, detects high-risk interactions, and triggers adaptive replanning when necessary.

Given an initial task plan $T = \{a_1, a_2, ..., a_n\}$, where each action $a_i$ corresponds to a specific interaction with an object or agent, our goal is to construct an adaptive spatio-semantic graph $G = (V, E)$, where $V$ represents environmental entities (e.g., objects, humans, robots), and $E$ encodes risk-aware and spatial relationships between them. The agent must update $T$ in response to high-risk edges $e_{ij} \in E$, ensuring safe task execution.

\subsection{Graph-Based Risk Representation}
We employ LLM-based semantic reasoning to annotate interactions between entities, assigning risk levels and  explainable hazard assessments. Specifically, given a scene with entities $\{v_1, v_2, ..., v_m\}$, an LLM generates structured risk annotations:
% \begin{equation}
%     \begin{split}
%         r(v_i, v_j) &= \text{LLM}(v_i, v_j) \\
%         & \text{where } r \in 
%         \{\text{low:0.25, medium:0.50, high:1.00}\}
%     \end{split}
% \end{equation}
\vspace{-5pt}
\begin{equation}
    \begin{aligned}
        r(v_i, v_j) &= \text{LLM}(v_i, v_j) \\
        r &\in \{\text{low:0.25, medium:0.50, high:1.00}\}
    \end{aligned}
\end{equation}
where $r(v_i, v_j)$ defines the inferred risk level between entities $v_i$ and $v_j$. Example risk annotations include:

\begin{tcolorbox}[colback=gray!10, colframe=gray!40, arc=2mm, boxrule=0.5pt]
\begin{lstlisting}[language=json]
{
    "type1": "Baby",
    "type2": "Kettle",
    "danger_level": "high",
    "risk_type": [
        "thermal",
        "physical",
        "water"
    ],
    "llm_reason": "A kettle, when in use,
    can become extremely hot and poses a risk 
    of burns or scalds to a baby. Additionally, 
    if a kettle is tipped over, it can lead to
    significant injury. The risk of water 
    spillage further increases the danger 
    as it can lead to slips or electrical 
    hazards if near power outlets."
}
\end{lstlisting}
\end{tcolorbox}

% \begin{itemize}
%     \item $\text{Child} \rightarrow \text{Stove}: \text{ High burn risk}$
%     \item $\text{Child} \rightarrow \text{Knife}: \text{ Potential cut hazard}$
% \end{itemize}

We construct a 	synthetic risk-aware dataset by embedding these risk relationships into AI2-THOR environments. We introduce randomized human-agent interactions (e.g., children, adults, pets) to ensure diverse safety-critical scenarios. Each sample is labeled with a 	danger score, computed as:
\begin{equation}
    S(v_i, v_j) = r(v_i, v_j) \times \text{SP}(v_i, v_j)
\end{equation}
where the SP refers to spatial proximity, which accounts for spatial scene configurations, and in experiment part we chose: 
% $max(1/distance,distance_threshold)$.
\begin{equation}
SP =
\begin{cases} 
\frac{1}{\text{distance}}, & \text{if } \text{distance} \leq \text{DT} \\ 
\frac{1}{\text{DT}}, & \text{otherwise}
\end{cases}
\end{equation}
where DT refers to the threshold to prevent SP from diverging.
% \vspace{-10pt}
\subsection{Graphormer Pretraining for Risk Detection}
We train a Graphormer\cite{graphormer} model to predict high-risk interactions based on the structured safety graph. Given an environment represented as a graph $\mathcal{G} = (\mathcal{V}, \mathcal{E})$ with node features $X \in \mathbb{R}^{|\mathcal{V}| \times d}$ and adjacency matrix $A \in \mathbb{R}^{|\mathcal{V}| \times |\mathcal{V}|}$, Graphormer computes risk-weighted attention scores via:
% \vspace{-5pt}
\begin{equation}
    H = \text{Graphormer}(X, A, E)
\end{equation}
% \vspace{-10pt}
where $H \in \mathbb{R}^{|\mathcal{V}| \times d}$ represents the learned risk embeddings, and $E \in \mathbb{R}^{|\mathcal{E}| \times d}$ encodes edge features. The model applies multi-head attention over graph structures:
% \vspace{-5pt}
\begin{equation}
    Z = \text{MultiHeadAttn}(Q, K, V) = \sum_{i=1}^{h} \alpha_i W_i V
\end{equation}
% \vspace{-5pt}
where queries, keys, and values are computed as:
\begin{equation}
    Q = W_Q H, \quad K = W_K H, \quad V = W_V H
\end{equation}
The attention weights $\alpha_i$ incorporate both structural and semantic information:
% \vspace{-5pt}
\begin{equation}
    \alpha_{ij} = \frac{\exp \left( \frac{(Q W_Q) (K W_K)^T}{\sqrt{d}} + b_{ij} \right)}{\sum_{k} \exp \left( \frac{(Q W_Q) (K W_K)^T}{\sqrt{d}} + b_{ik} \right)}
\end{equation}
% \vspace{-5pt}
where $b_{ij}$ represents edge encodings extracted from $E$. To address data imbalance—where most edges are non-hazardous—we employ a focal loss\cite{focalloss}:
% \vspace{-5pt}
\begin{equation}
    \mathcal{L} = - \sum_{(i,j) \in \mathcal{E}} w_{ij} (1 - p_{ij})^\gamma \log p_{ij}
\end{equation}
% \vspace{-5pt}
where $p_{ij}$ is the predicted probability of an edge being hazardous, $w_{ij}$ is a re-weighting factor prioritizing risky edges, and $\gamma$ adjusts the focus on hard-to-classify edges.
% \vspace{-5pt}

\subsection{Risk-Aware Task Replanning}
At runtime, Graphormer continuously evaluates the environment, identifying high-risk edges $e_{ij}$ in the safety graph. When a high-risk interaction is detected, the LLM is queried to generate a revised task plan:
\begin{equation}
    T' = \text{LLM}(T, G) \quad \text{subject to safety constraints.}
\end{equation}

The updated task plan incorporates safety-aware modifications, ensuring secure execution. An example of updated plan is as follows:

\begin{tcolorbox}[colback=gray!10, colframe=gray!40, arc=2mm, boxrule=0.5pt, breakable]
\textbf{Task:} Prepare a meal

\textbf{Description:}  Cook the meal.\\

\textbf{LLM Initial Plan:}  

0. Walk to kitchen  \\
1. Gather ingredients  \\
2. Start cooking  \\

\textbf{Graphormer Risk Detection:} Upon entering the kitchen, the system updates the environment state and detects a high-risk interaction:  

- High-risk edge detected: \\
Baby $\rightarrow$ Knife (Risk level: High)  \\
- Reason: A child is standing close to the knife, posing a potential injury hazard. \\ 

\textbf{Replanning Triggered:} The LLM generates a revised task sequence to mitigate the identified risk.  \\

\textbf{LLM Updated Plan:}  

0. Walk to kitchen  \\
1. Ensure child is in a safe location  \\
2. Secure knife in a designated area  \\
3. Gather ingredients  \\
4. Start cooking  \\
5. DONE  
\end{tcolorbox}

% This framework enables real-time adaptation, allowing robots to execute tasks in dynamic and safety-critical environments.

\begin{figure*}[htbp]
    \hfill
    \centering
    \resizebox{0.9\textwidth}{7.5cm}{
    \includegraphics[width=0.9\textwidth]{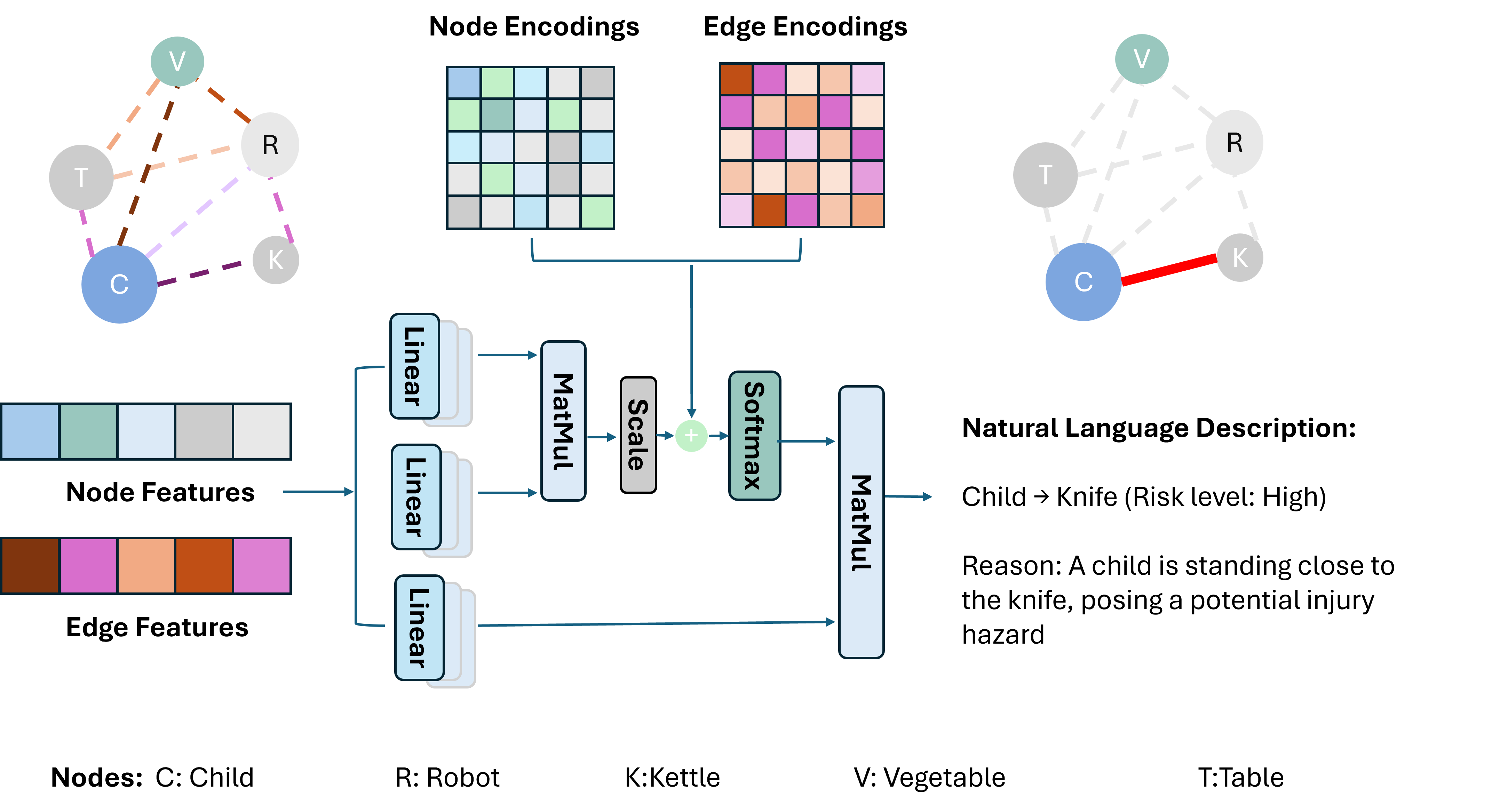}}
\caption{ Graphormer Overview. Our framework integrates Graphormer-based risk modeling with LLM-driven task planning to enable instantaneous safety adaptation. The system constructs a context-aware spatio-semantic safety graph from environmental observations, where high-risk interactions are identified using attention-weighted edge representations.The Graphormer will auto translates the dangerous edges into natural language.}

    \label{fig:overview}
\end{figure*}

\section{EXPERIMENTS}

\subsection{Risk Detection Experiments}
% In real-world robotics, ensuring safe task execution in dynamic environments is critical, particularly in shared spaces with human interactions. Our study focuses on risk-aware task planning in domestic environments, where a robotic agent must complete tasks while actively identifying and mitigating potential hazards. The key challenge lies in adapting the task sequence based on dynamically emerging risks, such as a child approaching a dangerous object or an environmental condition changing unpredictably. 

To evaluate our approach, we conduct experiments in AI2-THOR using a dataset of 120 diverse household environments, covering four common domestic settings: kitchens, living rooms, bedrooms, and bathrooms. To simulate real-world safety-critical scenarios, we manually introduce human agent nodes near hazardous objects (e.g., a child near a stove or a knife) in half of the scenes. The dataset is split into 90 training samples, 15 validation samples, and 15 test samples. We evaluate our method against two baseline approaches to analyze its performance in in-time, risk-aware task execution.

\subsubsection{Evaluation Metrics and Data Imbalance}
Our dataset exhibits significant class imbalance, with hazardous edges constituting only 1\% of the total edges. A naive classifier that predicts all edges as "safe" would trivially achieve 99\% accuracy while failing to identify critical safety risks.

\subsubsection{Comparison with Alternative Methods}

The results confirms that:

\begin{itemize}
    \item \textbf{Random Guess Baseline}: Performs no better than chance, with near-zero precision.
    \item \textbf{Rule-Based System}: Requires manual rule specification, which is inherently limited. No existing rule set comprehensively covers all indoor safety hazards, and manually defining exhaustive safety constraints is infeasible. In addition, static rule enforcement lacks adaptability, making it ineffective in continuous environments where unforeseen risks arise.
    \item \textbf{Our Model}: Achieves the best trade-off between hazard detection and task efficiency, with online risk perception to assess and adapt to environmental threats.
\end{itemize}

\begin{figure}[h]
    \centering
    \includegraphics[width=0.48\textwidth]{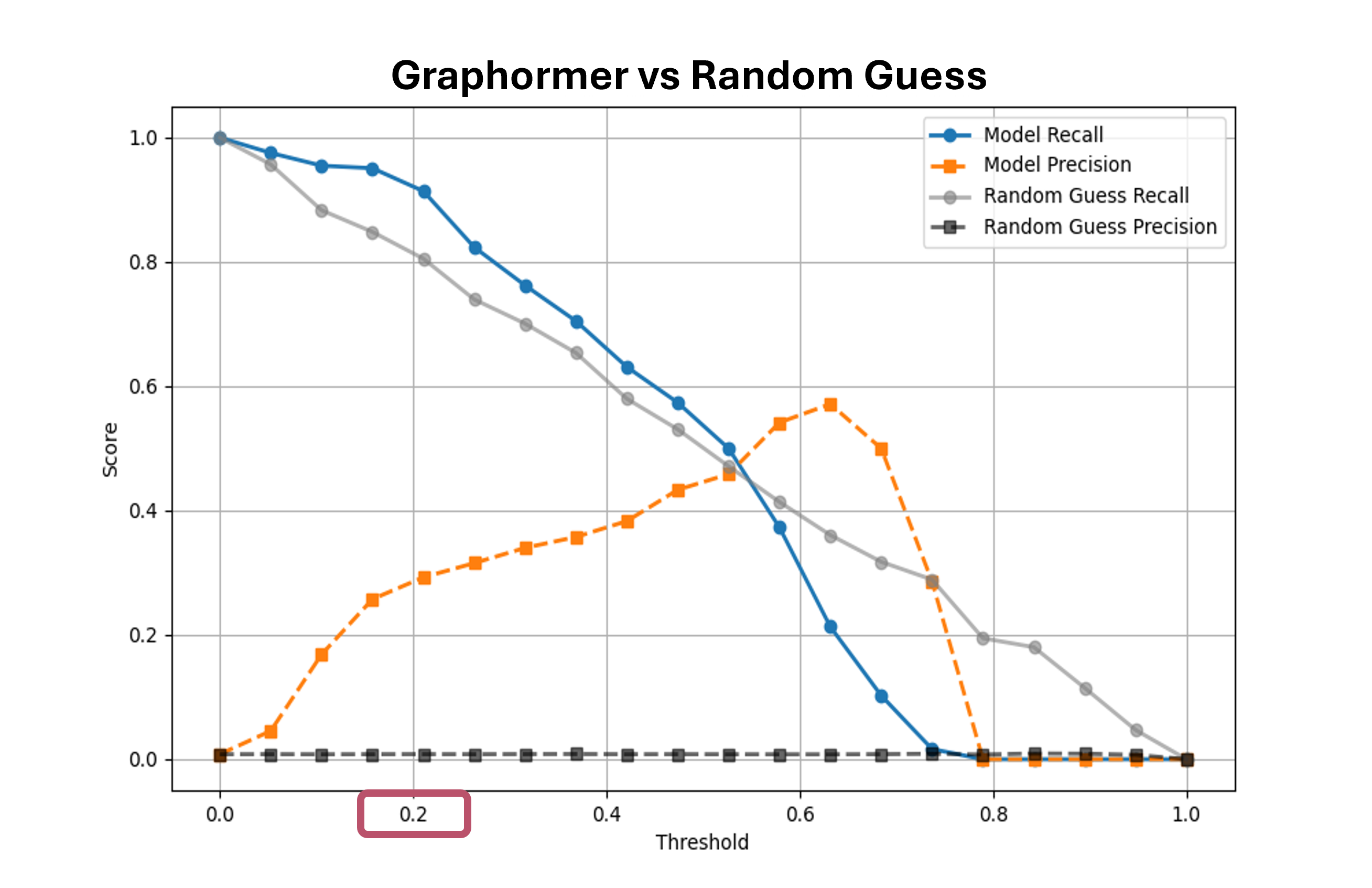}
\caption{Precision-Recall analysis of our method. Due to the severe class imbalance in our dataset, random guessing yields near-zero precision. In contrast, by setting a decision threshold of $0.21$, our method achieves a precision of 30\% while maintaining a recall above 90\%. This means that in a dataset with 10,000 edges, where only 100 are hazardous, our model identifies 300 edges as potentially dangerous, successfully capturing 90 out of the 100 true hazardous edges. This balance ensures that critical risks are detected while minimizing false alarms.}
\vspace{-15pt}
    \label{fig:recall}
\end{figure}

Figure \ref{fig:recall} illustrates the trade-off between Recall and Precision across different models.
To ensure a meaningful evaluation, we prioritize Recall as our primary metric, ensuring that hazardous interactions are detected. Specifically, we select a decision threshold=0.21 that yields:
\begin{itemize}
    \item \textbf{Recall:} 91.39\% (covering 90\% of hazardous edges)
    \item \textbf{Precision:} 29.27\% (filtering out 70\% of false positives)
\end{itemize}
This threshold ensures that safety-critical hazards are not overlooked while maintaining practical alert filtering.

\subsection{LLM-Guided Risk-Aware Task Planning Experiments}

\begin{table*}[t]
    \centering
    \renewcommand{\arraystretch}{1.3} 
    \setlength{\tabcolsep}{4pt}
    \caption{Task Planning Performance Across Different Tasks For Three Categories of Complexities}
    \begin{tabular}{l ccc ccc ccc}
        \toprule
        \multirow{2}{*}{\textbf{Model}} & \multicolumn{3}{c}{\textbf{Simple}} & \multicolumn{3}{c}{\textbf{Intermediate}} & \multicolumn{3}{c}{\textbf{Complex}} \\
        \cmidrule(lr){2-4} \cmidrule(lr){5-7} \cmidrule(lr){8-10}
        & TSR (\%) & SNR (\%) & RHS (\%) & TSR (\%) & SNR (\%) & RHS (\%) & TSR (\%) & SNR (\%) & RHS (\%) \\
        \midrule
        LLM-Only & 100 & 0 & 0 & 97.5 & 0 & 0 & 85.0 & 0 & 10 \\
        LLM-Safe-Prompting & 100 & 0 & 0 & 97.5 & 0 & 0 & 85.0 & 0 & 0 \\
        LLM + LTL & 95.0 & 100 & 90 & 90.0 & 100 & 80 & 60.0 & 100 & 40 \\
        % \rowcolor{gray!15} 
       \textbf{LLM + Graphormer (Ours)} & \textbf{100} & \textbf{100} & \textbf{100} & \textbf{97.5} & \textbf{100} & \textbf{95.0} & \textbf{92.5} & \textbf{100} & \textbf{95.0} \\
        \bottomrule
    \end{tabular}
    \label{tab:task_planning}
\end{table*}

To evaluate our proposed framework, we define 5 household activities categorized in three levels of complexities, including two spatial reorganization tasks(simple),  two object manipulation tasks(intermediate), and one cooking tasks(complex) .Each task requires no high-risk actions to be executed without proper intervention. We conducted real task planning experiments for these methods:
\begin{itemize}
    \item \textbf{LLM-Only (No Risk Awareness)}: The LLM receives the entire scene description and generates a task sequence without explicit risk perception.
    \item \textbf{Prompted LLM (Risk-Aware Prompting)}: The LLM is provided with a general instruction to consider safety risks but only receives the information of all objects without guidelines from environment.
    \item \textbf{LLM + Static Rules (LTL)}:
    The LLM is given a general instruction to consider safety risks and receives signals from the LTL module. In our experiments, we manually define rules to detect all hazards; however, this approach is not feasible in real-world settings.

    % The LLM is provided with a general instruction to consider safety risks and receives signals from the LTL module, we manually set rules that will detect all hazards in our experiments, which will not be capable in realistic settings.
    
    \item \textbf{LLM + Graphormer (Ours)}: Our model integrates risk-aware graph-based reasoning with LLM-based planning.
\end{itemize}
\subsubsection{Evaluation Metrics}
For additional analysis, we test our framework in AI2-THOR first 20 kitchen scenes with evaluation metrics:
\begin{itemize}
    \item \textbf{Task Success Rate (TSR)}: The percentage of successfully completed tasks.
    \item \textbf{Safety Notice Rate (SNR)}: The percentage of scenarios where a high-risk situation is noticed by agents.
    \item \textbf{Risk Handling Success (RHS)}: The percentage of cases where the system correctly identified and mitigated a potential hazard.
\end{itemize}
To further validate generalization capabilities, we introduce modified flexible human-agent interactions within AI2-THOR, simulating a child is near a dangerous object (e.g. knife). 
% Our system integrates an LLM-based task planner with a Graphormer-enhanced safety perception module. The framework consists of:
% \begin{itemize}
%     \item A 	Graphormer-based risk predictor that dynamically constructs a 	spatio-semantic safety graph, encoding interactions between objects and agents.
%     \item An 	LLM-based task planner that leverages safety insights to adjust the task execution sequence in real time.
%     \item A 	risk-aware task execution module that determines whether a replanning event is necessary, ensuring that hazardous interactions are mitigated.
% \end{itemize}
% We train Graphormer using a synthetic risk-aware dataset constructed from AI2-THOR, where diverse object-agent configurations are introduced to simulate realistic hazards. The LLM is responsible for reasoning over the graph structure and generating adaptive task plans.

% \subsubsection{Experiment Setup}

\subsubsection{Results and Analysis}
As shown in Table \ref{tab:task_planning}, our approach verifies superior risk detection and task modification capabilities compared to baselines. 

LLM-based planners fail to recognize dynamic risk relationships, even when prompted for safety. They may identify hazardous objects (e.g., knives) but lack structured perception to detect evolving threats (e.g., a child approaching). 

LTL-based methods achieve high accuracy in controlled settings due to manually predefined constraints, but real-world hazard enumeration is infeasible. Their static nature causes them to misclassify unforeseen risks as "safe," misleading the LLM.

Our method surpasses baselines in unstructured risk scenarios by leveraging a risk-aware safety graph, enabling proactive hazard detection and adaptive task planning for safer execution in dynamic environments.

\begin{figure}
    \centering
    \includegraphics[width=0.9\linewidth]{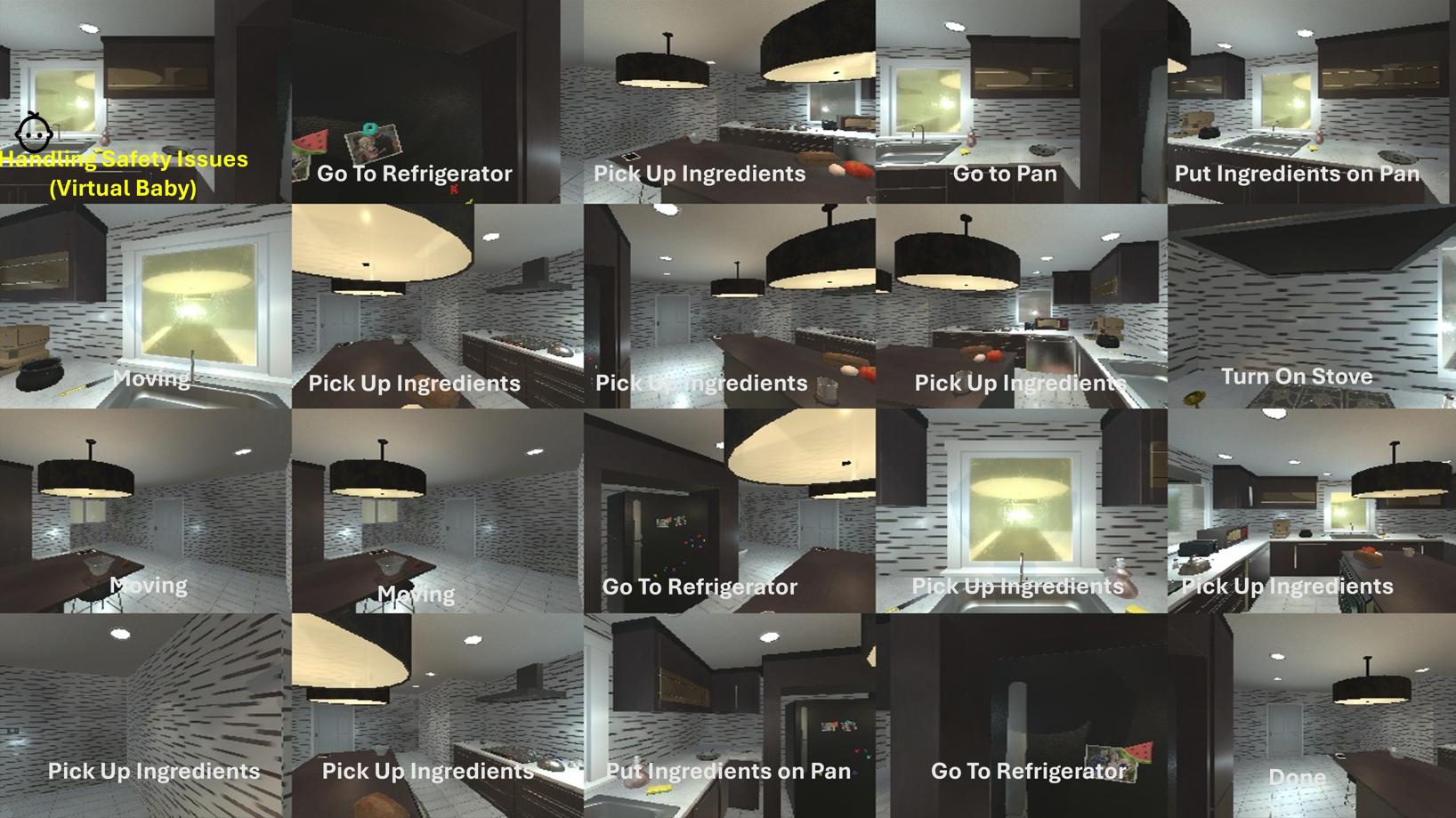}
    \caption{Stages of a complete cooking task in AI2-THOR (FloorPlan2) from the perspective of the executing agent. The task includes picking up ingredients and placing them into a pan. The first stage, \textit{"HandleSafetyIssue"} for the \textit{"Baby"}, is not natively supported in AI2-THOR; however, we implement a script-based check to determine whether the model actively resolves safety issues.}
\vspace{-15pt}
    % \caption{A whole cooking task stages for real-time environment in AI2-THOR at FloorPlan2 under the view of the executing agent, including picking up ingredient objects and put them into a pan object. The first stage in this task sequence is 
% "HandleSafetyIssue" for "Baby",
             % which is not support by native AI2-THOR but we check it in scripts to find if the model is resolving safety issues.}
    \label{fig:simulation}
\end{figure}

Fig. \ref{fig:simulation} illustrates task execution in AI2-THOR, highlighting our system's ability to adaptively modify task sequences based on emerging risks.

% While LLM-based planners exhibit general reasoning capabilities, they fail to recognize complex risk relationships, regardless of whether they are explicitly prompted to consider safety. For example, an LLM planner may acknowledge that a knife is dangerous but lacks the structured perception to detect that a child is approaching the object, leading to unsafe task execution.

% LTL-based approaches achieve high accuracy in our controlled experiments; however, this is largely due to the manual predefinition of all relevant safety constraints. In real-world scenarios, exhaustively enumerating all possible hazards is infeasible. Their constraints are inherently static—if an unforeseen risk emerges (e.g., an object not included in the predefined constraint set), the system defaults to classifying the scenario as "safe," potentially misleading the LLM into executing hazardous tasks.

% Our method consistently outperforms baselines in handling unstructured risk scenarios, exhibiting its robustness in domestic environments. The risk-aware safety graph enables proactive hazard detection by continuously modeling spatio-semantic relationships, ensuring safer and more reliable task execution.

% \subsubsection{Graphormer Risk Prediction Accuracy}
% We first evaluate Graphormer’s performance in predicting high-risk edges in the spatio-semantic safety graph. The results in Table \ref{tab:risk_prediction} demonstrate that our model significantly outperforms heuristic rule-based baselines in terms of precision and recall.

% \begin{table}[h]
%     \centering
%     \caption{Risk Prediction Performance of Graphormer(Replace with Real Data)}
%     \begin{tabular}{|l|c|c|}
%         \hline
%         \textbf{Model} & \textbf{Precision} & \textbf{Recall} \\
%         \hline
%         Rule-Based Heuristics & 74.2\% & 61.8\% \\
%         Graphormer (Ours) & \textbf{92.1\%} & \textbf{90.7\%} \\
%         \hline
%     \end{tabular}
%     \label{tab:risk_prediction}
% \end{table}

% \subsubsection{Adaptation to AI2-THOR Modifications}

% \section{Discussion}

% Our experiments demonstrate that integrating Graphormer with LLM-based task planning yields significant improvements in both risk detection accuracy and safety-aware decision-making. The ablation study confirms that LLMs alone struggle to capture fine-grained risk relations, whereas Graphormer-enhanced models excel in both static and dynamically changing environments.

% Future work will explore multi-modal risk integration, incorporating real-time sensory inputs such as visual perception and depth estimation to further enhance the adaptability of our system.

\section{CONCLUSION AND FUTURE WORK}

In this work, we introduced a graphormer-enhanced risk-aware task planning framework that integrates structured safety perception with LLM-driven decision-making for robotic agents in real-time environments. 
% Our approach constructs a spatio-semantic safety graph to model hazardous interactions, enabling immedeiate risk prediction and adaptive task planning. By exploiting Graphormer for structured risk reasoning and 	LLM for task adaptation, our system achieves superior safety and flexibility compared to existing rule-based and reinforcement learning baselines.
Through extensive evaluations in AI2-THOR, we validated the effectiveness of our method in 	three key aspects: 
\begin{itemize}
    \item 	Risk Prediction Accuracy: Our Graphormer model achieves high precision and recall in hazard identification.
    \item 	Risk-Aware Task Planning: Our LLM + Graphormer framework successfully adapts task sequences based on environmental risks, reducing unsafe actions while maintaining high task efficiency.
    \item 	Generalization in Real-time Environments: Our method consistently outperforms baselines in the rate of saftey notice and handling safety issues when tested in AI2-THOR settings.
\end{itemize}

% The qualitative analysis emphasizes the role of a risk-aware safety graph in mitigating unsafe interactions, proving its effectiveness in real-world evolving environments. Our study highlights the necessity of combining a Graphormer as the risk aware model with the LLM-driven planner, proving that neither component alone achieves optimal performance in safety-critical scenarios. 

While our framework demonstrates strong performance in risk-aware robotic task planning, several directions remain open for future exploration:
\begin{itemize}
\item Expanding the AI2-THOR environment with a broader set of interactive, safety-related objects to enhance risk-aware interactions. Specifically, we aim to release a systematically curated dataset to benchmark models on hazardous factor perception.
\item Extending our current static spatial risk perception to a temporal framework, enabling not just the recognition of hazards but also their prediction over time.
\end{itemize}

By addressing these challenges, we aim to further advance the intersection of LLM-driven task planning and graph-based safety modeling, paving the way for safer, more intelligent autonomous systems in unstructured environments.

\section*{Appendix}

% \subsection*{Time Consumption Analysis}

\begin{table}[h]
    \centering
        \caption{Comparison of Task Execution Times}
    \begin{tabular}{lcc}
        \toprule
        \textbf{Stage} & \textbf{Graphormer} (seconds) & \textbf{LTL} (seconds) \\
        \midrule
        Retrieve Object Information & 0.1552 & 0.1257 \\
        Build Environment Graph & 0.0185 & - \\
        Receive Safety Notice & 1.1926  & 0.0002 (1 rule) \\
        Generate Task Sequence & 1.4581 & 0.9931 \\
        Parse Task Sequence & 0.0000 & 0.0000 \\
        \bottomrule
    \end{tabular}

    \label{tab:time_comparison}
\end{table}

% \subsection*{Analysis and Conclusion}
Table \ref{tab:time_comparison} presents a comparison of task execution times for a single task between our method, Graphormer (loading model from disk), and LTL. It is important to emphasize that in the LTL setup, we used only one rule, whereas real-world applications require thousands of safety rules.As a result, the majority of time consumption is concentrated on waiting for the LLM to generate the task sequence, rather than on risk detection.

% From , it is evident that the primary time consumption in the Graphormer method stems from receiving the safety notice and generating the task sequence. In contrast, the LTL method requires significantly less time for safety verification due to the use of only one rule. However, in real-world applications, LTL requires thousands of safety rules, which would significantly increase its verification time.

% Thus, our method exhibits strong scalability in large-scale applications compared to LTL, as it avoids exponential increases in safety verification time due to rule growth. \textbf{Most of the time consumption remains focused on retrieving environmental information and waiting for the LLM's task generation response}, indicating that our approach remains efficient and scalable in real-world applications.

% 放点任务decription和prompt
% \section*{ACKNOWLEDGMENT}
% We thank the AI2-THOR community for providing a versatile simulation environment, and acknowledge insightful discussions with our research collaborators that helped shape this work. 

\bibliographystyle{ieeetr} 
\bibliography{graphrobotics}

\end{document}